\pdfoutput=1

\documentclass[11pt]{article}

\usepackage[final]{acl}

\usepackage{times}
\usepackage{latexsym}

\usepackage[T1]{fontenc}

\usepackage[utf8]{inputenc}

\usepackage{microtype}

\usepackage{inconsolata}

\usepackage{amsmath}
\usepackage{amssymb}
\usepackage{mathtools}
\usepackage{amsthm}
\usepackage{graphicx}
\usepackage{booktabs} %
\usepackage{multirow}
\usepackage{multicol}
\usepackage{xcolor}
\usepackage{tikz}
\usepackage{pgfplots}
\pgfplotsset{compat=1.3}
\usepackage{subcaption}

\newcommand{\Lc}{\mathcal{L}}

\newcommand{\Sc}{\mathcal{S}}
\newcommand{\Tc}{\mathcal{T}}

\newcommand{\Rb}{\mathbb{R}}

\newcommand{\ev}{\mathbf{e}}

\newcommand{\yv}{\mathbf{y}}

\newcommand{\Av}{\mathbf{A}}
\newcommand{\Bv}{\mathbf{B}}

\newcommand{\Dv}{\mathbf{D}}

\newcommand{\Wv}{\mathbf{W}}
\newcommand{\Xv}{\mathbf{X}}

\title{OWSM-CTC: An Open Encoder-Only Speech Foundation Model for Speech Recognition, Translation, and Language Identification}

\author{Yifan Peng \\
  Carnegie Mellon University \\
  \texttt{yifanpen@andrew.cmu.edu} \\\And
  Yui Sudo \\
  Honda Research Institute Japan \\
  \texttt{yui.sudo@jp.honda-ri.com} \\\AND
  Muhammad Shakeel \\
  Honda Research Institute Japan \\
  \texttt{shakeel.muhammad@jp.honda-ri.com} \\\And
  Shinji Watanabe \\
  Carnegie Mellon University \\
  \texttt{swatanab@andrew.cmu.edu} \\
  }

\begin{document}
\maketitle
\begin{abstract}
There has been an increasing interest in large speech models that can perform multiple tasks in a single model. Such models usually adopt an encoder-decoder or decoder-only architecture due to their popularity and good performance in many domains. 
However, autoregressive models can be slower during inference compared to non-autoregressive models and also have potential risks of hallucination.
Though prior studies observed promising results of non-autoregressive models for certain tasks at small scales, it remains unclear if they can be scaled to speech-to-text generation in diverse languages and tasks.
Inspired by the Open Whisper-style Speech Model (OWSM) project, we propose OWSM-CTC, a novel encoder-only speech foundation model based on Connectionist Temporal Classification (CTC). It is trained on 180k hours of public audio data for multilingual automatic speech recognition (ASR), speech translation (ST), and language identification (LID).
Compared to encoder-decoder OWSM, our OWSM-CTC achieves competitive results on ASR and up to 24\% relative improvement on ST, while it is more robust and 3 to 4 times faster for inference.
OWSM-CTC also improves the long-form ASR result with 20x speed-up.
We will publicly release our code, pre-trained model, and training logs to promote open science in speech foundation models.\footnote{\url{https://github.com/espnet/espnet}}
\end{abstract}

\section{Introduction}
\label{sec:introduction}

\begin{figure}[t]
     \begin{subfigure}[b]{\linewidth}
          \centering
          \begin{tikzpicture}
	\begin{axis}[
		xlabel=Speed-up ($\rightarrow$),
		ylabel=WER ($\leftarrow$),
		xtick={0, 1, 2, 3, 4},
            ytick={8, 10, ..., 14},
    	xmin=0.5,
            xmax=4.5,
            ymin=7,
            ymax=13,
            ylabel shift = -3pt,
            xlabel shift = -4pt,
            label style={font=\footnotesize},
            ticklabel style={font=\scriptsize},
            width=\linewidth,
            height=0.4\linewidth,
	    every axis plot/.append style={thick},
            legend cell align={left},
            legend columns=1,
            legend style={at={(0,1)},anchor=north west,nodes={scale=0.55, transform shape}}
		]

        \addplot[color=blue, solid, mark=square, mark options={scale=0.8, solid}] coordinates {
(1, 7.7)
(2.97, 12.1)
	}; 
        \addlegendentry{OWSM v3.1 series};

        \node [above] at (axis cs:  1,  7.7) {\textcolor{blue}{\scriptsize medium}};
        \node [below] at (axis cs:  2.97, 12.1) {\textcolor{blue}{\scriptsize base}};
        
	\addplot[color=red, solid, mark=star, mark options={scale=2, solid}] coordinates {
(3.63, 7.9)
	}; 
        \addlegendentry{OWSM-CTC (ours)};

	\end{axis}
\end{tikzpicture}
          \vskip -0.05in
          \caption{English speech recognition}
     \end{subfigure}
     \begin{subfigure}[b]{\linewidth}
          \centering
          \begin{tikzpicture}
	\begin{axis}[
		xlabel=Speed-up ($\rightarrow$),
		ylabel=BLEU ($\rightarrow$),
		xtick={0, 1, 2, 3, 4},
            ytick={5, 10, ..., 35},
    	xmin=0.5,
            xmax=4.5,
            ymin=7,
            ymax=35,
            ylabel shift = -3pt,
            xlabel shift = -4pt,
            label style={font=\footnotesize},
            ticklabel style={font=\scriptsize},
            width=\linewidth,
            height=0.4\linewidth,
	    every axis plot/.append style={thick},
            legend cell align={left},
            legend columns=1,
            legend style={at={(0,1)},anchor=north west,nodes={scale=0.55, transform shape}}
		]

        \addplot[color=blue, solid, mark=square, mark options={scale=0.8, solid}] coordinates {
(1, 20.2)
(2.78, 9.6)
	}; 
        \addlegendentry{OWSM v3.1 series};

        \node [below] at (axis cs:  1, 20.2) {\textcolor{blue}{\scriptsize medium}};
        \node [above] at (axis cs:  2.78, 9.6) {\textcolor{blue}{\scriptsize base}};
        
	\addplot[color=red, solid, mark=star, mark options={scale=2, solid}] coordinates {
(3.35, 25.1)
	}; 
        \addlegendentry{OWSM-CTC (ours)};

	\end{axis}
\end{tikzpicture}
          \vskip -0.05in
          \caption{X-to-En speech translation}
     \end{subfigure}
     \begin{subfigure}[b]{\linewidth}
          \centering
          \begin{tikzpicture}
	\begin{axis}[
		xlabel=Speed-up ($\rightarrow$),
		ylabel=BLEU ($\rightarrow$),
		xtick={0, 1, 2, 3, 4, 5},
            ytick={5, 10, ..., 30},
    	xmin=0.5,
            xmax=4.5,
            ymin=3,
            ymax=25,
            ylabel shift = -3pt,
            xlabel shift = -4pt,
            label style={font=\footnotesize},
            ticklabel style={font=\scriptsize},
            width=\linewidth,
            height=0.4\linewidth,
	    every axis plot/.append style={thick},
            legend cell align={left},
            legend columns=1,
            legend style={at={(0,1)},anchor=north west,nodes={scale=0.55, transform shape}}
		]

        \addplot[color=blue, solid, mark=square, mark options={scale=0.8, solid}] coordinates {
(1, 14.1)
(2.39, 5.5)
	}; 
        \addlegendentry{OWSM v3.1 series};

        \node [below] at (axis cs:  1, 14.1) {\textcolor{blue}{\scriptsize medium}};
        \node [above] at (axis cs:  2.39, 5.5) {\textcolor{blue}{\scriptsize base}};
        
	\addplot[color=red, solid, mark=star, mark options={scale=2, solid}] coordinates {
(4.2, 16.0)
	}; 
        \addlegendentry{OWSM-CTC (ours)};

	\end{axis}
\end{tikzpicture}
          \vskip -0.05in
          \caption{En-to-X speech translation}
     \end{subfigure}
     \caption{
     Performance vs. speed for encoder-decoder OWSM v3.1 and our encoder-only OWSM-CTC.
     }
     \label{fig:performance-speed}
\end{figure}
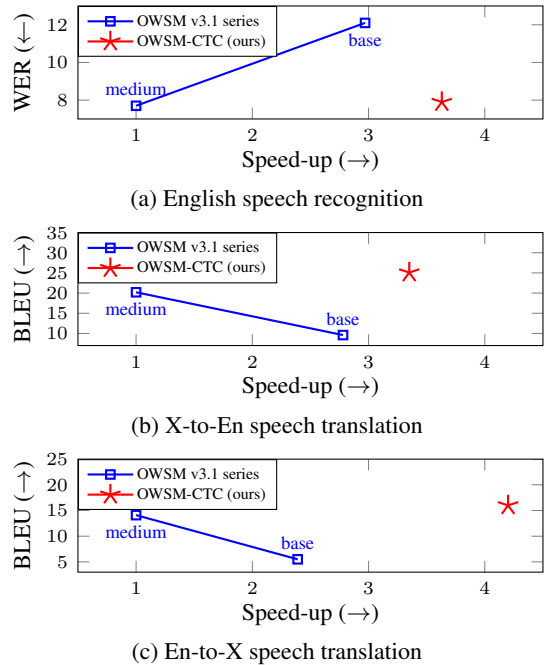

The great success of large language models (LLMs)~\citep{gpt4,llama2,palm2} has sparked a growing interest in developing foundation models in various modalities. Recent studies have explored different approaches towards multilingual and multi-tasking speech foundation models~\citep{whisper, google-usm, meta-mms, audiopalm, seamless, owsm-asru23}.
OpenAI Whisper~\citep{whisper} is a series of Transformer encoder-decoder models trained on 680k hours of proprietary labeled audio. Whisper achieves strong results in multilingual automatic speech recognition (ASR), any-to-English speech translation (ST), and spoken language identification (LID). Although it shows the effectiveness of large-scale (weakly) supervised pre-training, the full development pipeline, including training data details, is not publicly accessible. 
Recent works have developed Open Whisper-style Speech Models (OWSM)~\citep{owsm-asru23, owsm-v31} with the aim of reproducing Whisper-style training using public data and open-source toolkits. However, Whisper and OWSM adopt the encoder-decoder architecture, which generates text tokens given speech in an autoregressive manner. They might hallucinate during inference, and the speed can be slow. Other models with decoder-only architectures, like AudioPaLM~\citep{audiopalm} and VioLA~\citep{viola}, could suffer from the same issues due to autoregressive decoding.

Another type of work like Google USM~\citep{google-usm} and Meta MMS~\citep{meta-mms} uses non-autoregressive models with Connectionist Temporal Classification (CTC)~\cite{ctc}, but these CTC-based models are designed for ASR only. Prior studies have also achieved promising results of CTC models for ST only, but they mainly focus on specific language pairs at much smaller scales~\citep{orthros, chuang-etal-2021-investigating, xu-etal-2023-ctc}. Some of them employ additional decoders~\citep{orthros, yan-etal-2023-ctc} or cross-attention layers~\citep{xu-etal-2023-ctc}, making the model more complicated.

A natural question now arises: \emph{Can we build a non-autoregressive encoder-only model for speech-to-text generation in diverse languages and multiple tasks like Whisper/OWSM?}
This research problem has become increasingly important in the era of LLMs because large-scale pre-trained speech encoders can serve as an adapter between the speech and text modalities~\citep{ltu, google-slm}, providing a promising avenue towards general-purpose multi-modal foundation models~\citep{gemini}.

In this work, we propose OWSM-CTC, a novel encoder-only speech foundation model based on multi-task self-conditioned CTC to imitate OWSM's multilingual ASR, any-to-any ST, and LID functionalities.
Following previous encoder-decoder OWSM v3.1 models~\citep{owsm-v31},
we train a 1B OWSM-CTC model using 180k hours of public data covering 151 languages.
Extensive evaluations show that our OWSM-CTC exhibits strong performance and efficiency. Compared to the 1B OWSM v3.1 medium model, OWSM-CTC achieves comparable performance for ASR and superior performance for various ST directions (up to 24\% relative improvement) while being more robust and showing 3 to 4 times inference speed-up.
OWSM-CTC also improves the WER for long-form ASR and can be 20 times faster due to batched parallel decoding.
OWSM-CTC further outperforms the other baseline models on LID.
Our code, pre-trained model weights, and training logs will be publicly released to facilitate the development of large speech models.

\section{Related Work}

\subsection{Speech foundation models}

\noindent \textbf{Attention-based encoder-decoder.}
OpenAI Whisper~\citep{whisper} adopts the standard Transformer encoder-decoder architecture~\citep{transformer} and scales the training data to 680k hours of proprietary labeled audio.\footnote{Their latest large-v3 version uses 1M hours of labeled audio and 4M hours of pseudo-labeled audio.} However, the complete pipeline for model development, including training data details and training code, is not publicly available. A recent project, OWSM, aims to reproduce Whisper-style training using public data and open-source toolkits to promote transparency and open science in this field~\citep{owsm-asru23}. The latest OWSM v3.1 models~\citep{owsm-v31} employ E-Branchformer~\citep{ebranchformer} as the encoder and Transformer as the decoder, which are trained with a joint ASR CTC loss~\citep{joint-ctc-attn}. Although OWSM has promising results using public corpora, it still follows the encoder-decoder architecture, which can be slow and unstable at inference time.

\noindent \textbf{Decoder-only.}
Several studies employ decoder-only models for speech-to-text tasks. AudioPaLM~\citep{audiopalm} extends the textual PaLM-2~\citep{palm2} to support speech understanding and generation tasks including ASR and ST. DOTA~\citep{dota} is a decoder-only Transformer model trained on 93k hours of public English ASR data, but it does not support other languages or ST. Decoder-only models face the same slowness and robustness issues as encoder-decoder due to autoregressive decoding.

\noindent \textbf{CTC or Transducer.}
Another line of research proposes to utilize CTC~\citep{ctc} or Transducer~\citep{rnn-t} for ASR. Google USM~\citep{google-usm} provides generic ASR models that are first pre-trained on 12M hours of unlabeled audio and then fine-tuned on proprietary labeled data with CTC or Transducer. 
Meta MMS~\citep{meta-mms} pre-trains a wav2vec 2.0 model~\citep{wav2vec2} on massively multilingual data and then fine-tunes it with CTC on labeled ASR data covering over 1k languages. These models employ CTC only for ASR. In our OWSM-CTC, we propose a single CTC-based encoder-only model for ASR, ST, and LID. Our supported tasks are more similar to Whisper-style models.

\subsection{Efficient speech models}

\noindent \textbf{Model compression.}
Various algorithms have been utilized to compress speech models, including knowledge distillation~\citep{distilhubert, fithubert, dphubert, distil-whisper}, pruning~\citep{parp, structured-pruning}, quantization~\citep{quantize-ssl, usm-lite}, and dynamic module execution~\citep{hubert-ee, i3d, pmlr-v202-strimel23a}. These methods are typically applied to pre-trained models and are thus orthogonal to this work. In the future, we will apply compression to further improve efficiency. 

\noindent \textbf{Efficient architectures.}
Better network architectures can also improve efficiency, including attention with linear complexity~\citep{longformer, linformer, efficient-transformer-survey} and sequence length reduction~\citep{efficientconformer, squeezeformer, nawrot-etal-2023-efficient, fastconformer}. In this work, we do not modify the attention but use larger downsampling in the convolution module to reduce the sequence length. More details are in Appendix~\ref{app:model-arch} and \ref{app:effect-downsampling}.

\begin{figure*}[tbp!]
    \centering
    \includegraphics[width=0.85\textwidth]{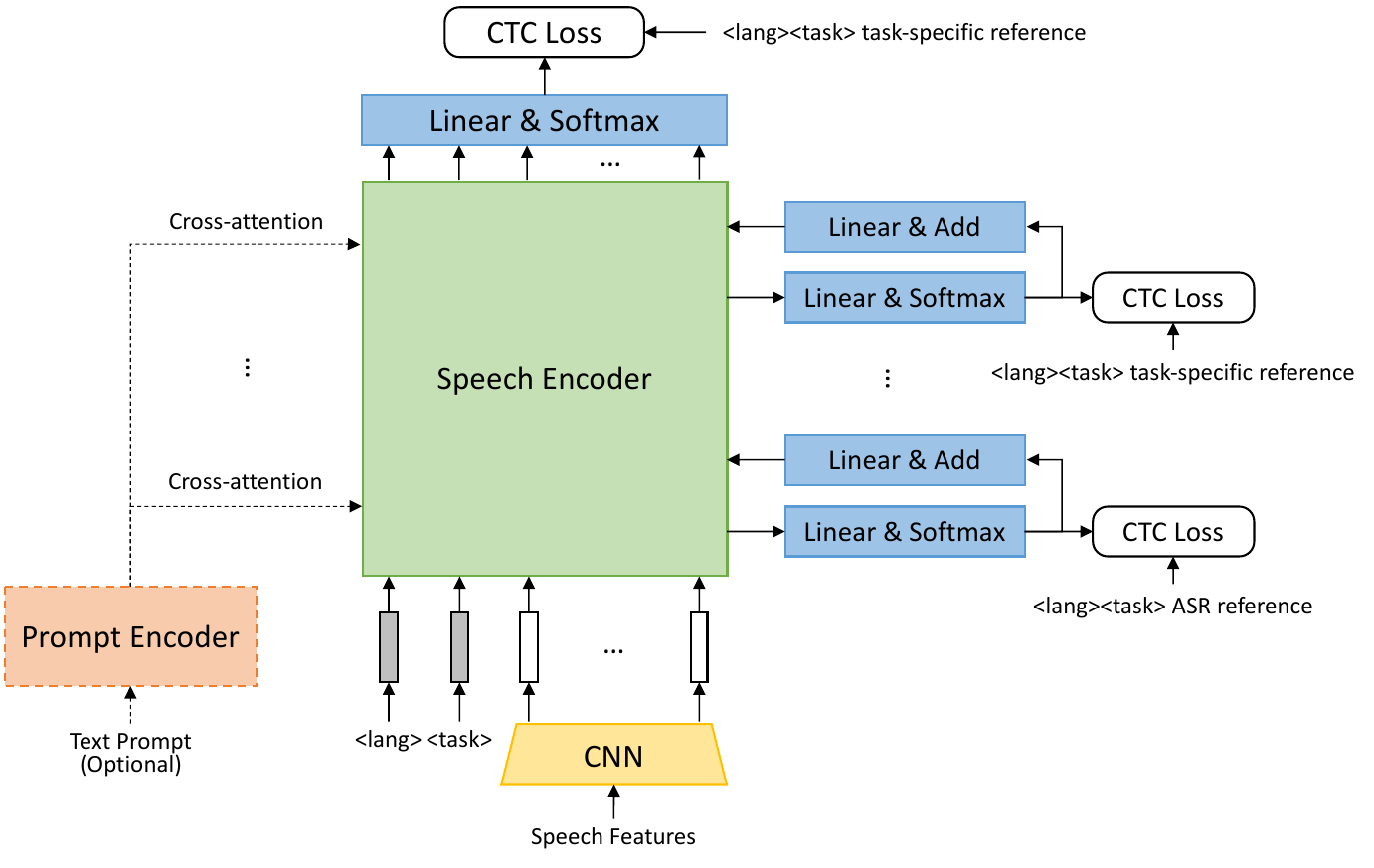}
    \vskip -0.13in
    \caption{Architecture of our OWSM-CTC. For an input audio, it predicts a language token along with ASR or ST text tokens depending on the task specifier.
    An optional text prompt can be provided, which mimics Whisper.
    }
    \vskip -0.13in
    \label{fig:model-arch}
\end{figure*}

\subsection{CTC-based speech models}
\label{subsec:ctc-background}
Non-autoregressive models have a faster inference speed than their autoregressive counterparts due to parallel decoding. They have been utilized in machine translation~\citep{gu2018nonautoregressive, ghazvininejad-etal-2019-mask, survey-nar-nmt}, ASR~\citep{nar-asr, mask-ctc, limit-of-nar-asr, chi-etal-2021-align, interctc, self-conditioned-ctc}, and ST~\citep{orthros, chuang-etal-2021-investigating, xu-etal-2023-ctc}. 

CTC is originally proposed to label sequences without explicit segmentation~\citep{ctc}. CTC-based ASR models learn a monotonic alignment between speech features and text tokens. With parallel greedy decoding, they are much faster than autoregressive models.
However, the accuracy of CTC is generally inferior due to the conditional independence assumption between output tokens.
To address this issue, Intermediate CTC (InterCTC)~\citep{interctc} calculates additional CTC losses using intermediate representations from the encoder. Self-conditioned CTC~\cite{self-conditioned-ctc} further extends InterCTC by adding back predictions of intermediate CTC layers to the subsequent encoder. These approaches have shown to be highly effective in speech-to-text generation tasks without a decoder~\citep{nar-study}.

Although CTC assumes a monotonic alignment between input and output, it can be used for ST with the reordering capability of self-attention~\citep{orthros, chuang-etal-2021-investigating}.

Conventional CTC models are typically designed for a specific task or language. It remains under-explored whether such approaches can be scaled to multilingual and multi-task scenarios.
This work proposes a novel encoder-only speech foundation model based on multi-task self-conditioned CTC. This single model performs well in multilingual ASR, ST, and LID.

\section{OWSM-CTC}

\subsection{Overall architecture}

\autoref{fig:model-arch} shows the architecture of OWSM-CTC. Its main component is a speech encoder, which takes speech features as input and predicts the spoken language as well as the ASR or ST hypothesis using CTC.
To mimic Whisper-style models that condition text generation on an optional text prompt~\citep{whisper, owsm-asru23, owsm-v31}, we employ a separate Transformer encoder to process the prompt and inject the output to the main model through cross-attention. Then, the model can potentially attend to the text prompt when generating text.

\subsection{Speech encoder}
\label{subsec:speech-encoder}

For an input waveform, we first extract log Mel filterbanks and then apply a 2D convolution module to downsample the feature sequence along the time dimension. Let $\Xv_{\text{speech}} \in \Rb^{T\times d}$ be the downsampled feature sequence of length $T$ and feature size $d$. To specify the language and task, we prepend two special tokens to the sequence:
\begin{align}
    \Xv = \text{concat}(\ev_{\text{lang}}, \ev_{\text{task}}, \Xv_{\text{speech}}),
\end{align}
where $\text{concat}(\cdot)$ is concatenation along time and $\ev_{\text{lang}}, \ev_{\text{task}} \in \Rb^{1\times d}$ are embeddings of special tokens \texttt{<lang>} and \texttt{<task>}, respectively. $\Xv$ now has shape $(T+2)\times d$.
If the spoken language is known, the true language token will be used as input. Otherwise, a special token \texttt{<nolang>} denoting ``unknown language'' will be used. During training, we randomly replace the true language with \texttt{<nolang>} according to probability 0.5 so that either can be used for inference.
The task token is \texttt{<asr>} for speech recognition and \texttt{<st\_lang>} for translation to a target language.

Next, we add sinusoidal positional embeddings to $\Xv$, and apply a stack of $N$ encoder layers:
\begin{align}
    \Xv^{(0)} &= \Xv + \text{PosEmb}(\Xv),\\
    \Xv^{(l)} &= \text{SpeechEnc}^{(l)}(\Xv^{(l-1)}), \label{eq:vanilla-enc}
\end{align}
where $l$ is a layer index from 1 to $N$, $\text{PosEmb}(\cdot)$ generates positional embeddings, and $\text{SpeechEnc}^{(l)}(\cdot)$ is the $l$-th encoder layer. The encoder is E-Branchformer~\citep{ebranchformer}, an enhanced version of Branchformer~\citep{branchformer}, which shows excellent performance across a wide range of benchmarks~\citep{ebf-vs-conformer}.

We compute the CTC loss using the final encoder output $\Xv^{(N)}$ and an augmented reference $\yv_{\text{task}}$. To create this reference, we simply preprend \texttt{<lang>} and \texttt{<task>} to the original groundtruth text of the desired task. Hence, the model will learn to predict the language token in addition to ASR or ST text tokens. This CTC loss is denoted as follows:
\begin{align}
\resizebox{0.88\linewidth}{!}{
 $   \Lc^{(N)} = -\log P_{\text{CTC}}(\yv_{\text{task}} \mid \text{softmax}( \Xv^{(N)} \Wv_1)), \label{eq:ctcloss-last} $
}
\end{align}
where $\Wv_1 \in \Rb^{d\times V}$ is a linear layer and $V$ is the size of the CTC vocabulary.

As discussed in Section~\ref{subsec:ctc-background}, we apply self-conditioned CTC~\citep{self-conditioned-ctc} at intermediate layers $\Sc \subseteq \{1,\ldots,N-1\}$ to alleviate the conditional independence assumption of CTC. For any layer $s \in \Sc$, \autoref{eq:vanilla-enc} is replaced by the following operations:
\begin{align}
    \Av^{(s)} &= \text{SpeechEnc}^{(s)}(\Xv^{(s-1)}), \label{eq:enc-int}\\
    \Bv^{(s)} &= \text{softmax}(\Av^{(s)} \Wv_1),\\
    \Xv^{(s)} &= \Av^{(s)} + \Bv^{(s)} \Wv_2,
\end{align}
where $\Wv_2 \in \Rb^{V\times d}$ is a linear layer. The intermediate CTC loss at layer $s$ is defined as follows:
\begin{align}
    \Lc^{(s)} = -\log P_{\text{CTC}}(\yv^{(s)} \mid \Bv^{(s)}), \label{eq:ctcloss-int}
\end{align}
where $\yv^{(s)}$ is the augmented reference at layer $s$. 
Similar to $\yv_{\text{task}}$ in \autoref{eq:ctcloss-last}, we prepend the language and task tokens to the original groundtruth text. 
Note that the choice of the reference text depends on the task. 
If the task for the current input is ASR, we simply use the ASR transcript to create $\yv^{(s)}$ for all $s$, which is consistent with conventional ASR models.
However, if the task is ST, we empirically find that the model cannot converge if we use the translated text as the reference at all intermediate layers $\Sc$ (see Appendix~\ref{app:choice-ctc-ref} for discussions). Therefore, as shown in \autoref{fig:model-arch}, we utilize the ASR transcript at the first $N_{\text{ASR}}$ layers and the ST text at the remaining $N_{\text{ST}}$ layers, where $N_{\text{ASR}} + N_{\text{ST}} = |\Sc| \leq N-1$. This design mimics a cascaded system that first performs ASR and then ST, but our entire model is optimized jointly and trained from scratch.
In other words, the first $N_{\text{ASR}}$ CTC layers always perform ASR regardless of the task token (named ``ASR-only CTC''), whereas the other CTC layers are multi-tasking - they can perform ASR or ST according to the task token (named ``task-specific or task-dependent CTC'').

The overall training loss is an average of the loss terms defined in \autoref{eq:ctcloss-last} and \autoref{eq:ctcloss-int}:
\begin{align}
    \Lc_{\text{total}} = \frac{1}{1 + |\Sc|} \left(\Lc^{(N)} + \sum_{s\in\Sc} \Lc^{(s)} \right).
\end{align}

\subsection{Prompt encoder}
\label{subsec:prompt-encoder}

Whisper-style models generate text conditioned on an optional text prompt~\citep{whisper, owsm-asru23, owsm-v31}. During training, this prompt is simply the previous sentence in the same audio recording. During inference, it can be provided by the user to potentially adjust the output.
For encoder-decoder models like Whisper, the text prompt is a prefix to the autoregressive decoder. For our encoder-only model, we leverage a separate Transformer encoder to process the prompt and inject it to the speech encoder through cross-attention. If no prompt is provided, a special token \texttt{<na>} will be used. Let $\Xv_{\text{prompt}} \in \Rb^{T^\prime \times d^\prime}$ be the output of the prompt encoder. We insert a cross-attention layer at a subset of layers $\Tc \subseteq \{1,\ldots,N\}$ of the speech encoder. For any $t \in \Tc$, the original $\text{SpeechEnc}^{(t)}(\cdot)$ in \autoref{eq:vanilla-enc} or \autoref{eq:enc-int} becomes $\text{SpeechEncCA}^{(t)}(\cdot, \cdot)$:
\begin{align}
    &\Dv^{(t)} = \text{SpeechEnc}^{(t)}(\Xv^{(t-1)}), \\
    &\text{SpeechEncCA}^{(t)}(\Xv^{(t-1)},\Xv_{\text{prompt}}) = \notag\\
    &~~~~\Dv^{(t)} + \text{CrossAtt}(\Dv^{(t)}, \Xv_{\text{prompt}}, \Xv_{\text{prompt}}),
\end{align}
where $\text{CrossAtt}(\cdot,\cdot,\cdot)$ is a cross-attention layer with three arguments: query, key, and value.

Our training data is a mixture of public ASR and ST datasets. Some of them provide unsegmented long audio, but the others only release segmented short audio. At training time, if a sample does not have a previous sentence, we will use \texttt{<na>}. Otherwise, we use either \texttt{<na>} or the previous sentence as the prompt according to 0.5 probability. Section~\ref{subsec:effect-text-prompt} shows that OWSM-CTC can leverage the prompt's information when necessary.

\begingroup
\begin{table}[tb]
  \centering
  \resizebox {\linewidth} {!} {
  \begin{tabular}{l|cccc}
    \toprule
    & Params & Time shift & Training data & GPU hours \\
    \midrule
    \multicolumn{5}{l}{\textbf{Whisper (encoder-decoder)}~\citep{whisper}}\\
    base & 74M & 20ms & 680k hours & unknown\\
    small & 244M & 20ms & 680k hours & unknown\\
    medium & 769M & 20ms & 680k hours & unknown\\
    large-v2 & 1550M & 20ms & 680k hours & unknown\\
    \midrule
    \multicolumn{5}{l}{\textbf{OWSM v3.1 (encoder-decoder)}~\citep{owsm-v31}}\\
    base & 101M  & 40ms & 180k hours & 2.3k\\
    medium & 1.02B & 40ms & 180k hours & 24.6k\\
    \midrule
    \multicolumn{5}{l}{\textbf{OWSM-CTC (ours)}}\\
    medium & 1.01B & 80ms & 180k hours & 19.2k\\
    \bottomrule
  \end{tabular}
  }
  \vskip -0.05in
  \caption{Summary of model size, training data, and training cost measured on an NVIDIA A100 GPU (40GB).
  }
  \vskip -0.1in
  \label{tab:mainbody-exp-setups}
\end{table}
\endgroup

\section{Experiments}

\subsection{Experimental setups}

\autoref{tab:mainbody-exp-setups} is a brief summary of model size, training data, and training cost.

\noindent \textbf{Data format.}
Our training data is prepared using scripts publicly released by OWSM v3.1~\citep{owsm-v31}. It is a mixture of more than 25 public ASR and ST corpora covering 151 languages and various translation directions. The total audio duration is 180k hours. To create long-form data, consecutive utterances from the same audio recording are concatenated to a duration of no more than 30 seconds. The input audio to the model is always padded to a fixed length of 30 seconds. Appendix~\ref{app:training-data} and \autoref{tab:training-data} present the training data statistics.
The original Whisper-style data contains the start and end timestamps for each utterance. These timestamp tokens are predicted along with normal text tokens during the autoregressive decoding. In OWSM-CTC, we do not include any explicit timestamps since the time-aligned hypothesis can be obtained by forced alignment if desired.

\noindent \textbf{Model architecture.}
Our speech encoder is a 27-layer E-Branchformer with a hidden size of 1024 and 16 attention heads. Four intermediate layers (6, 12, 15, and 21) are used for self-conditioned CTC. The first three are ASR only, while the others are task-specific.
The prompt encoder is a 4-layer Transformer with a hidden size of 512 and 8 attention heads. It is injected into the speech encoder at every third layer.
The total model size is 1.01B, which matches the size of the encoder-decoder OWSM v3.1 medium (1.02B).
More details about the architecture are in Appendix~\ref{app:model-arch} (see \autoref{tab:model-arch}).

\noindent \textbf{Implementation.}
We implement OWSM-CTC in ESPnet~\citep{espnet} based on PyTorch~\citep{pytorch}. FlashAttention~\citep{flash-attn} is used to improve training efficiency, but it is not used for inference.
The batch size per GPU is 4, and 64 NVIDIA A100 GPUs (40GB) are used with distributed data parallel. The total training time is approximately 300 hours.
For optimization, we employ the Adam optimizer~\citep{adam-opt} with the piece-wise linear learning rate schedule~\citep{owsm-v31}. The peak learning rate is 2e-4.
Other training hyperparameters can be found in Appendix~\ref{app:training-hyper} (see \autoref{tab:training-hyper}).

\noindent \textbf{Evaluation.}
We fairly compare our encoder-only OWSM-CTC with the previously released encoder-decoder OWSM v3.1 models~\citep{owsm-v31} since they are trained on the same data. We also show the results of Whisper under the same decoding setup for reference, but we note that they are not comparable with ours due to completely different training data.
By default, short-form audio without any text prompt is used, but we also evaluate the long-form ASR performance in Section~\ref{subsec:long-asr} and investigate the effect of text prompts in Section~\ref{subsec:effect-text-prompt}.

\begingroup
\begin{table}[tb]
  \centering
  \resizebox {0.9\linewidth} {!} {
  \begin{tabular}{l|c}
    \toprule
    & Accuracy \% ($\uparrow$) \\
    \midrule
    \multicolumn{2}{l}{\textbf{Whisper (encoder-decoder)}~\citep{whisper}}\\
    \quad base & 47.6\\
    \quad small & 53.1\\
    \quad medium & 54.8\\
    \midrule
    \midrule
    \multicolumn{2}{l}{\textbf{OWSM v3 (encoder-decoder)}~\citep{owsm-asru23}}\\
    \quad medium & 81.4\\
    \midrule
    \multicolumn{2}{l}{\textbf{OWSM v3.1 (encoder-decoder)}~\citep{owsm-v31}}\\
    \quad base & 41.9\\
    \quad medium & 75.6\\
    \midrule
    \multicolumn{2}{l}{\textbf{OWSM-CTC (ours)}}\\
    \quad medium & \textbf{87.6}\\
    \bottomrule
  \end{tabular}
  }
  \vskip -0.05in
  \caption{Spoken LID results on the FLEURS test set.
  }
  \vskip -0.05in
  \label{tab:lid}
\end{table}
\endgroup

\subsection{Language identification}

\autoref{tab:lid} presents the LID results on the FLEURS test set~\citep{fleurs}. Our OWSM-CTC achieves a top-1 accuracy of 87.6\%, outperforming the other encoder-decoder models by a large margin.
This is likely because spoken LID requires a powerful encoder to extract useful information from the input audio. Our encoder-only model is especially suitable for this type of task.

\begingroup
\setlength{\tabcolsep}{3pt}
\begin{table}[tb!]
  \centering
  \resizebox {\linewidth} {!} {
  \begin{tabular}{l|ccccccccc|cc}
    \toprule
    & \rotatebox{90}{CommonVoice en} & \rotatebox{90}{FLEURS en} & \rotatebox{90}{LibriSpeech test-clean} & \rotatebox{90}{LibriSpeech test-other} & \rotatebox{90}{MLS en} & \rotatebox{90}{Switchboard eval2000} & \rotatebox{90}{TEDLIUM} & \rotatebox{90}{VoxPopuli en} & \rotatebox{90}{WSJ eval92} & \rotatebox{90}{Average WER ($\downarrow$)} & \rotatebox{90}{Speed-up ($\uparrow$)} \\
    \midrule
    \multicolumn{12}{l}{\textbf{Whisper (encoder-decoder)}~\citep{whisper}}\\
    base & 25.2 & 12.4 & 5.1 & 12.0 & 13.4 & 25.7 & 6.3 & 10.2 & 5.0 & 12.8 & 2.40x\\
    small & 15.7 & 9.6 & 3.3 & 7.7 & 9.1 & 22.2 & \textbf{4.6} & 8.5 & 4.3 & 9.4 & 1.46x\\
    medium & \textbf{11.9} & \textbf{6.4} & 2.8 & 6.5 & 10.2 & 19.4 & 5.1 & \textbf{7.6} & \textbf{2.9} & 8.1 & 0.76x\\
    \textcolor{gray!75}{large-v2} & \textcolor{gray!75}{10.5} & \textcolor{gray!75}{6.0} & \textcolor{gray!75}{4.1} & \textcolor{gray!75}{6.1} & \textcolor{gray!75}{7.7} & \textcolor{gray!75}{24.0} & \textcolor{gray!75}{6.0} & \textcolor{gray!75}{7.1} & \textcolor{gray!75}{3.3} & \textcolor{gray!75}{8.3} & \textcolor{gray!75}{0.55x} \\
    \midrule
    \midrule
    \multicolumn{12}{l}{\textbf{OWSM v3.1 (encoder-decoder)}~\citep{owsm-v31}}\\
    base & 21.5 & 14.8 & 3.6 & 9.1 & 12.0 & 22.9 & 7.8 & 12.0 & 5.3 & 12.1 &  2.97x\\
    medium & 12.6 & 9.0 & \textbf{2.4} & \textbf{5.0} & \textbf{7.1} & \textbf{16.3} & 5.1 & 8.4 & 3.5 & \textbf{7.7} & 1.00x\\
    \textcolor{gray!75}{+ beam 5} & \textcolor{gray!75}{11.7} & \textcolor{gray!75}{8.5} & \textcolor{gray!75}{2.7} & \textcolor{gray!75}{5.3} & \textcolor{gray!75}{6.6} & \textcolor{gray!75}{15.5} & \textcolor{gray!75}{5.1} & \textcolor{gray!75}{8.5} & \textcolor{gray!75}{3.4} & \textcolor{gray!75}{7.5} & \textcolor{gray!75}{0.06x}\\
    \midrule
    \multicolumn{12}{l}{\textbf{OWSM-CTC (ours)}}\\
    medium & \underline{12.1} & 9.9 & \textbf{2.4} & 5.2 & 7.3 & 16.9 & \underline{4.9} & 8.6 & 4.2 & 7.9 & \underline{\textbf{3.63x}}\\
    \bottomrule
  \end{tabular}
  }
  \vskip -0.05in
  \caption{WER \% ($\downarrow$) of English ASR. Speed-up ($\uparrow$) is based on average decoding time. Whisper is trained on 438k hours of English audio, whereas OWSM v3.1 and our OWSM-CTC are trained on only 73k hours. 
  Results of Whisper large-v2 and OWSM v3.1 medium with beam search are shown in \textcolor{gray!75}{gray}, which are not comparable with the others due to different model sizes or decoding configurations.
  \textbf{Bold}: the best result. \underline{Underlined}: OWSM-CTC outperforms OWSM v3.1 medium.
  }
  \label{tab:en-asr}
\end{table}
\endgroup

\begingroup
\setlength{\tabcolsep}{2.5pt}
\begin{table}[tb!]
  \centering
  \resizebox {\linewidth} {!} {
  \begin{tabular}{l|ccccccccccc|c}
    \toprule
    & \rotatebox{90}{MLS es} & \rotatebox{90}{MLS fr} & \rotatebox{90}{MLS de} & \rotatebox{90}{MLS nl} & \rotatebox{90}{MLS it} & \rotatebox{90}{MLS pt} & \rotatebox{90}{MLS pl} & \rotatebox{90}{AISHELL-1 (zh)} & \rotatebox{90}{KsponSpeech clean (ko)} & \rotatebox{90}{KsponSpeech other (ko)} & \rotatebox{90}{ReazonSpeech (ja)} & \rotatebox{90}{Average Error Rate ($\downarrow$)} \\
    \midrule
    \midrule
    data size & 11.1 & 9.8 & 13.3 & 2.1 & 2.6 & 8.6 & 4.3 & 23.4 & 8.0 & 8.0 & 7.1 \\
    \midrule
    \multicolumn{12}{l}{\textbf{Whisper (encoder-decoder)}~\citep{whisper}}\\
    base & 14.5 & 25.2 & 19.9 & 30.9 & 32.9 & 23.5 & 25.2 & 39.1 & 27.0 & 22.9 & 54.1 & 28.7\\
    small & 9.1 & 13.6 & 11.5 & 18.2 & 21.3 & 13.8 & 12.5 & 25.1 & 24.0 & 15.4 & 32.5 & 17.9\\
    medium & \textbf{6.1} & \textbf{9.7} & \textbf{8.1} & \textbf{12.2} & \textbf{15.6} & \textbf{8.9} & \textbf{6.8} & 15.7 & 17.6 & \textbf{12.8} & 25.3 & \textbf{12.6}\\
    \textcolor{gray!75}{large-v2} & \textcolor{gray!75}{4.8} &	\textcolor{gray!75}{7.0} &	\textcolor{gray!75}{6.3}	& \textcolor{gray!75}{9.7} &	\textcolor{gray!75}{13.2} &	\textcolor{gray!75}{6.6} &	\textcolor{gray!75}{5.5} &	\textcolor{gray!75}{18.3} &	\textcolor{gray!75}{20.0} &	\textcolor{gray!75}{13.1} &	\textcolor{gray!75}{26.8} &	\textcolor{gray!75}{11.9} \\
    \midrule
    \midrule
    data size & 2.0 & 2.5 & 3.7 & 1.7 & 0.7 & 0.3 & 0.3 & 16.3 & 1.0 & 1.0 & 18.9 \\
    \midrule
    \multicolumn{12}{l}{\textbf{OWSM v3.1 (encoder-decoder)}~\citep{owsm-v31}}\\
    base & 18.5 & 24.2 & 18.7 & 28.6 & 33.7 & 44.9 & 49.7 & 12.2 & 23.8 & 26.1 & 11.2 & 26.5\\
    medium & 9.0 & 12.1 & 10.8 & 18.1 & 20.2 & 21.6 & 25.2 & \textbf{6.4} & 16.7 & 18.9 & \textbf{7.9} & 15.2\\
    \textcolor{gray!75}{+ beam 5} & \textcolor{gray!75}{8.6} &	\textcolor{gray!75}{11.2} &	\textcolor{gray!75}{10.2} &	\textcolor{gray!75}{17.2} &	\textcolor{gray!75}{19.1} &	\textcolor{gray!75}{19.4} &	\textcolor{gray!75}{23.4} &	\textcolor{gray!75}{5.9} &	\textcolor{gray!75}{15.0} &	\textcolor{gray!75}{17.0} &	\textcolor{gray!75}{7.8} &	\textcolor{gray!75}{14.1} \\
    \midrule
    \multicolumn{12}{l}{\textbf{OWSM-CTC (ours)}}\\
    medium & 10.3 & 12.9 & 11.9 & 20.4 & 22.1 & 23.5 & 31.6 & \textbf{6.4} & \underline{\textbf{14.8}} & \underline{16.5} & 8.1 & 16.2\\
    \bottomrule
  \end{tabular}
  }
  \caption{Multilingual ASR results. CER~\% ($\downarrow$) is shown for Chinese (zh), Korean (ko) and Japanese (ja), while WER~\% ($\downarrow$) is shown for the others. Data sizes are in thousand hours. 
  Results of Whisper large-v2 and OWSM v3.1 medium with beam search are shown in \textcolor{gray!75}{gray}, which are not comparable with the others. \textbf{Bold}: the best result. \underline{Underlined}: OWSM-CTC outperforms OWSM v3.1 medium.
  }
  \vskip -0.15in
  \label{tab:multilingual-asr}
\end{table}
\endgroup

\subsection{Speech recognition}

\autoref{tab:en-asr} presents word error rates (WERs) on nine English ASR test sets. Following \citet{owsm-asru23, owsm-v31}, we leverage greedy decoding and apply the Whisper English text normalizer before scoring.\footnote{We also report the results of Whisper large-v2 and OWSM v3.1 medium with beam search in \textcolor{gray!75}{gray} for reference, but they are not comparable with the others due to different model sizes or decoding configurations. This applies to other tables as well.}
We record the average decoding time across all English test sets on an NVIDIA A40 GPU and calculate the relative speed-up.
Results show that our non-autoregressive OWSM-CTC generally has comparable WERs with the autoregressive OWSM v3.1 medium (average: 7.9 vs. 7.7), both of which have 1B parameters. 
However, OWSM-CTC achieves 3.63x speed-up due to parallel decoding. Notably, OWSM-CTC is even faster than OWSM v3.1 base, which has only 100M parameters, and our WERs are much lower (average: 7.9 vs. 12.1).
Compared to Whisper models trained on significantly more data, our OWSM-CTC is still competitive in many cases, and our inference is much faster.
These results demonstrate that OWSM-CTC achieves an excellent trade-off between recognition accuracy and inference efficiency.

\autoref{tab:multilingual-asr} shows the results of multilingual ASR. We perform greedy decoding and apply the Whisper basic text normalizer before scoring. Our OWSM-CTC is slightly worse than OWSM v3.1 in terms of the average WER/CER (16.2 vs. 15.2). For European languages in MLS~\citep{mls}, OWSM-CTC generally falls behind.
But for East Asian languages like Chinese, Japanese, and Korean, OWSM-CTC is on par with or better than OWSM v3.1 medium.
This difference might be related to the training data size and tokenization.

\begingroup
\setlength{\tabcolsep}{4.5pt}
\begin{table}[tb]
  \centering
  \resizebox {\linewidth} {!} {
  \begin{tabular}{l|cccc|cc}
    \toprule
    Src Lang. & de & es & fr & ca & Ave. ($\uparrow$) & Speed-up ($\uparrow$) \\
    \midrule
    \midrule
    data size & 4.3 & 6.7 & 4.5 & 0.2 \\
    \midrule
    \multicolumn{7}{l}{\textbf{Whisper (encoder-decoder)}~\citep{whisper}}\\
    base & 11.0	& 18.9 &	13.2 &	9.9 &	13.3 & 1.84x\\
    small & 23.9 &	31.8 &	26.1 &	21.4 &	25.8 & 1.54x\\
    medium & \textbf{32.0}	& \textbf{37.3} &	\textbf{33.4} &	\textbf{28.8} &	\textbf{32.9} & 0.84x\\
    \textcolor{gray!75}{large-v2} & \textcolor{gray!75}{35.2} &	\textcolor{gray!75}{39.7} &	\textcolor{gray!75}{35.7} &	\textcolor{gray!75}{31.2} &	\textcolor{gray!75}{35.5} & \textcolor{gray!75}{0.48x} \\
    \midrule
    \midrule
    data size & 0.2 & 0.1 & 0.3 & 0.1 \\
    \midrule
    \multicolumn{7}{l}{\textbf{OWSM v3.1 (encoder-decoder)}~\citep{owsm-v31}}\\
    base & 7.1 & 10.3 &	11.5 &	9.4 &	9.6 & 2.78x\\
    medium & 16.7	& 22.3 &	22.8 &	18.8 &	20.2 & 1.00x\\
    \textcolor{gray!75}{+ beam 5} & \textcolor{gray!75}{18.2} &	\textcolor{gray!75}{24.5} &	\textcolor{gray!75}{24.4} &	\textcolor{gray!75}{21.1} &	\textcolor{gray!75}{22.1}	& \textcolor{gray!75}{0.05x}\\
    \midrule
    \multicolumn{7}{l}{\textbf{OWSM-CTC (ours)}}\\
    medium & \underline{20.7}	& \underline{27.9} &	\underline{27.5} &	\underline{24.2} &	\underline{25.1} & \underline{\textbf{3.35x}}\\
    \bottomrule
  \end{tabular}
  }
  \vskip -0.05in
  \caption{BLEU ($\uparrow$) of X-to-En ST on CoVoST-2. Data sizes are in thousand hours. 
  Results of Whisper large-v2 and OWSM v3.1 medium with beam search are shown in \textcolor{gray!75}{gray}, which are not comparable with the others. \textbf{Bold}: the best result. \underline{Underlined}: OWSM-CTC outperforms OWSM v3.1 medium.
  }
  \label{tab:st-x-to-en}
\end{table}
\endgroup

\begingroup
\setlength{\tabcolsep}{3pt}
\begin{table*}[tb]
  \centering
  \resizebox {\linewidth} {!} {
  \begin{tabular}{l|ccccccccccccccc|cc}
    \toprule
    Tgt Lang. & de & ca & zh & fa & et & mn & tr & ar & sv & lv & sl & ta & ja & id & cy & Ave. ($\uparrow$) & Speed-up ($\uparrow$) \\
    \midrule
    data size & 14.0 & 0.4 & 13.7 & 0.8 & 0.4 & 0.4 & 0.9 & 0.9 & 0.4 & 0.4 & 0.4 & 0.4 & 1.0 & 0.4 & 0.4 & - & -\\
    \midrule
    \multicolumn{18}{l}{\textbf{OWSM v3.1 (encoder-decoder)}~\citep{owsm-v31}}\\
    base & 15.8 & 8.3 &	13.0 &	3.3 &	3.1 &	1.6 &	2.0 &	1.7 &	8.7 &	2.3 &	1.3 &	0.0 &	10.6 &	6.1 &	5.0 &	5.5 & 2.39x\\
    medium & 26.3 &	20.4 &	29.7 &	\textbf{10.2} &	9.6 &	5.8 &	7.8 &	7.2 &	20.8 &	8.4 &	11.0 &	0.1 &	\textbf{21.1} &	17.2 &	16.3 &	14.1 & 1.00x\\
    \textcolor{gray!75}{+ beam 5} & \textcolor{gray!75}{27.3} &	\textcolor{gray!75}{22.5} &	\textcolor{gray!75}{31.3} &	\textcolor{gray!75}{11.1} &	\textcolor{gray!75}{11.1} &	\textcolor{gray!75}{6.9} &	\textcolor{gray!75}{9.1} &	\textcolor{gray!75}{8.4} &	\textcolor{gray!75}{22.3} &	\textcolor{gray!75}{9.9} &	\textcolor{gray!75}{12.7} &	\textcolor{gray!75}{0.1} &	\textcolor{gray!75}{22.3} &	\textcolor{gray!75}{19.7} &	\textcolor{gray!75}{17.9} &	\textcolor{gray!75}{15.5} & \textcolor{gray!75}{0.05x}\\
    \midrule
    \multicolumn{18}{l}{\textbf{OWSM-CTC (ours)}}\\
    medium & \textbf{26.7}	& \textbf{24.0} &	\textbf{32.9} &	9.9 &	\textbf{11.4} &	\textbf{6.2} &	\textbf{7.9} &	\textbf{8.3} &	\textbf{24.5} &	\textbf{10.0} &	\textbf{14.2} &	0.1 &	20.4 &	\textbf{22.6} &	\textbf{20.6} &	\textbf{16.0} & \textbf{4.20x}\\
    p-value & 0.006 & 0.001 &	0.001 &	0.001 &	0.001 &	0.001 &	0.145 &	0.001 &	0.001 &	0.001 &	0.001 &	0.031 &	0.001 &	0.001 &	0.001 & - & -\\
    \bottomrule
  \end{tabular}
  }
  \vskip -0.05in
  \caption{BLEU ($\uparrow$) of En-to-X ST on CoVoST-2. Data sizes are in thousand hours. Note that Whisper does not support En-to-X translation. The p-values are computed by comparing OWSM-CTC against OWSM v3.1 medium using the Paired Significance Test in SacreBLEU~\citep{SacreBLEU}.
  Results of OWSM v3.1 medium with beam search are shown in \textcolor{gray!75}{gray}, which are not comparable with the others.
  }
  \vskip -0.1in
  \label{tab:st-en-to-x}
\end{table*}
\endgroup

\subsection{Speech translation}
\label{subsec:st-results}

We evaluate ST on CoVoST-2 test sets~\citep{covost2}. By default, we perform greedy decoding and calculate BLEU scores in true case with punctuation.\footnote{Results in lowercase without punctuation can be found in Appendix~\ref{app:more-st-results}, which are consistent with previous OWSM work~\citep{owsm-v31}.} For X-to-En translation, we follow OWSM~v3.1~\citep{owsm-v31} to report results of directions where the training data size is over 100 hours. For the other low-resource directions, both OWSM~v3.1 and our OWSM-CTC do not work in general. For En-to-X translation, we report all 15 directions.
We calculate the speed-up based on the average decoding time on an NIVIDA A40 GPU.

\autoref{tab:st-x-to-en} shows the X-to-En results. Notably, our encoder-only OWSM-CTC consistently outperforms the encoder-decoder OWSM v3.1 by a large margin. The average BLEU score is improved from 20.2 to 25.1 (24\% relatively). We also achieve 3.35x speed-up for inference.

\autoref{tab:st-en-to-x} presents En-to-X results. Whisper does not support these directions. Our OWSM-CTC achieves superior performance than OWSM v3.1 in 12  of 15 translation directions and most of them are statistically significant. The average BLEU is improved from 14.1 to 16.0 (13\% relatively), and the inference speed-up is 4.20 times.

We have the following observations from the ST results:
(1) Our non-autoregressive OWSM-CTC generally achieves 3 to 4 times speed-up compared to the encoder-decoder baseline, which is consistent with ASR.
(2) OWSM-CTC even improves the ST performance sometimes by a large margin. One reason is that the autoregressive model suffers from hallucination and error propagation, while the non-autoregressive model is more stable.
(3) The BLEU improvement of X-to-En is larger than that of En-to-X, likely because: (i) the OWSM training set contains lots of English ASR data and OWSM-CTC might obtain strong capability of generating English text; (ii) X-to-En has fewer training data than En-to-X, and the encoder-decoder model may need a sufficient amount of training data to achieve good performance for translation.

Our findings reveal that large-scale CTC-based models are also promising for ST in various language pairs, which is consistent with prior investigations at smaller scales~\citep{yan-etal-2023-ctc}.

\begingroup
\begin{table}[tb]
  \centering
  \resizebox {0.92\linewidth} {!} {
  \begin{tabular}{lc|cc}
    \toprule
    & Context Length &  WER \% ($\downarrow$) & Speed-up ($\uparrow$) \\
    \midrule
    \multicolumn{4}{l}{\textbf{Whisper (encoder-decoder)}~\citep{whisper}}\\
    base & - & 5.3 & 1.40x\\
    small & - & 4.4 & 1.62x\\
    medium & - & \textbf{3.8} & 0.86x\\
    \midrule
    \midrule
    \multicolumn{4}{l}{\textbf{OWSM v3.1 (encoder-decoder)}~\citep{owsm-v31}}\\
    base & - & 9.6 & 1.40x\\
    medium & - & 5.7 & 1.00x\\
    \midrule
    \multicolumn{4}{l}{\textbf{OWSM-CTC (ours)}}\\
    \multirow{4}{*}{medium} & 2s & \underline{5.4} & \underline{\textbf{22.40x}}\\
    & 4s & \underline{5.2} & \underline{19.35x}\\
    & 6s & \underline{5.2} & \underline{16.07x}\\
    & 8s & \underline{5.2} & \underline{12.09x}\\
    \bottomrule
  \end{tabular}
  }
  \vskip -0.05in
  \caption{Long-form ASR results on the TEDLIUM~\citep{tedlium3} test set which consists of 11 audio recordings ranging from 6 to 27 minutes.
  \textbf{Bold}: the best result. \underline{Underlined}: OWSM-CTC outperforms OWSM v3.1 medium.
  }
  \label{tab:long-asr}
\end{table}
\endgroup

\subsection{Long-form speech recognition}
\label{subsec:long-asr}

For long-form ASR, a model takes as input an unsegmented audio recording of arbitrary length and generates the entire transcription without explicit voice activity detection.
Whisper and encoder-decoder OWSM can predict start and end timestamps of each utterance within a fixed-length segment. Those timestamps are used to shift the recognition window for chunk-wise long-form ASR. However, this chunk-wise recognition is a sequential process because the location of the next chunk depends on the predicted timestamp in the current chunk.\footnote{The decoding process might be parallelized if token-level timestamps are available. However, it remains an open problem to derive accurate token-level timestamps from an attention-based encoder-decoder model without extra training.} 
By contrast, our OWSM-CTC performs chunk-wise recognition in a fully parallel manner. We first split the entire audio into overlapped chunks of 30s, where the overlapped region serves as the left and right context.\footnote{We follow this tutorial for long-form ASR with CTC: \url{https://github.com/NVIDIA/NeMo/blob/main/tutorials/asr/Streaming_ASR.ipynb}} We then perform CTC greedy decoding on batched chunks. The batch size is 32 on a single NVIDIA A40 GPU (48GB).
\autoref{tab:long-asr} shows the WER and speed-up with different context lengths. Our OWSM-CTC achieves lower WERs than the encoder-decoder OWSM v3.1, while being approximately 20 times faster due to the batched parallel decoding. OWSM-CTC is also robust to different context lengths.
These observations indicate that CTC-based non-autoregressive models perform very well for long-form ASR, which is consistent with prior findings~\citep{longform-asr}.

\begingroup
\setlength{\tabcolsep}{2pt}
\begin{table}[t!]
  \centering
  \resizebox {\linewidth} {!} {
  \begin{tabular}{lcccccc}
    \toprule
     & GigaSpeech &	LS-clean &	LS-other &	SWBD &	TEDLIUM &	AISHELL \\
    \midrule
    w/o prev & 11.80 & 2.42 & 5.22 & 16.92 & 4.95 & 6.37\\
    w/ prev & \textbf{11.23} & \textbf{2.38} &	\textbf{5.10} &	\textbf{16.70} &	\textbf{4.55} &	\textbf{6.25} \\
    \midrule
    p-value & <0.001 & 0.19 & 0.007 & <0.001 & <0.001 &	<0.001\\
    \bottomrule
  \end{tabular}
  }
  \vskip -0.05in
  \caption{Using the previous sentence as a text prompt improves the ASR WER/CER of OWSM-CTC.
  }
  \vskip -0.1in
  \label{tab:effect-prompt}
\end{table}
\endgroup

\subsection{Effect of text prompt}
\label{subsec:effect-text-prompt}

As described in \autoref{fig:model-arch} and Section~\ref{subsec:prompt-encoder}, OWSM-CTC can take an additional text prompt as input which might change the output. During training, either a special token \texttt{<na>} or the previous sentence in the same audio is used as the prompt according to a probability of 0.5, which follows the setup of Whisper and OWSM.
To verify that OWSM-CTC can utilize information from the prompt when necessary, we perform greedy decoding on several test sets with the previous sentence in the dataset as a prompt.
As shown in \autoref{tab:effect-prompt}, using the previous sentence reduces the error rates. The p-values are computed using the Matched Pair Sentence Segment method.\footnote{\url{https://github.com/usnistgov/SCTK}}
\autoref{app:example-prompt} provides an example where the previous sentence also affects the output text style.

\begingroup
\begin{table}[tb]
  \centering
  \resizebox {0.92\linewidth} {!} {
  \begin{tabular}{l|ccc}
    \toprule
    Input length & 5s & 10s & 20s \\
    \midrule
    \multicolumn{4}{l}{\textbf{Whisper (encoder-decoder)}~\citep{whisper}}\\
    large-v3 & Fjell & Fusilet & Rekordverk\\
    \midrule
    \multicolumn{4}{l}{\textbf{OWSM v3.1 (encoder-decoder)}~\citep{owsm-v31}}\\
    medium & thank you & thank you & (Applause)\\
    \midrule
    \multicolumn{4}{l}{\textbf{OWSM-CTC (ours)}}\\
    medium & . & ( & ( )\\
    \bottomrule
  \end{tabular}
  }
  \vskip -0.05in
  \caption{ASR outputs with random noise as input.
  }
  \vskip -0.15in
  \label{tab:robustness-random-input}
\end{table}
\endgroup

\subsection{Robustness}

To investigate the robustness, we first consider random noise as input. \autoref{tab:robustness-random-input} shows the ASR outputs generated by three models. Encoder-decoder models, including Whisper and OWSM v3.1, tend to generate some texts that look meaningful, while our OWSM-CTC generates fewer tokens, which are mostly punctuation marks that do not actually have meaning. 

Another typical issue of autoregressive decoding is that the generation might fall into repetitions of a few characters or words. \autoref{tab:robustness-example} in \autoref{app:robustness} presents two examples from ASR and ST, respectively. Our non-autoregressive model is more robust in such cases.
To quantitatively measure this type of error, we consider a hypothesis as a failure if it contains any character-level $\theta$-gram ($\theta=1,2,\dots,\theta_\text{max}$) that consecutively occurs for at least $\delta$ times. \autoref{tab:robustness-failures} shows the number of failures in all ST test sets with different thresholds. We can see that the encoder-decoder OWSM v3.1 medium fails many times even with beam search, while our OWSM-CTC has almost no failures.

\begingroup
\setlength{\tabcolsep}{8pt}
\begin{table}[tb]
  \centering
  \resizebox {\linewidth} {!} {
  \begin{tabular}{cclr}
    \toprule
    $\theta_\text{max}$ & $\delta$ & Model & \#Failures ($\downarrow$)\\
    \midrule
    \multirow{3}{*}{10} & \multirow{3}{*}{5} & OWSM v3.1 & 2448\\
    && OWSM v3.1 (beam 5) & 630\\
    && OWSM-CTC (ours) & 3\\
    \midrule
    \multirow{3}{*}{20} & \multirow{3}{*}{5} & OWSM v3.1 & 2537\\
    && OWSM v3.1 (beam 5) & 672\\
    && OWSM-CTC (ours) & 3\\
    \midrule
    \multirow{3}{*}{20} & \multirow{3}{*}{6} & OWSM v3.1 & 1985 \\
    && OWSM v3.1 (beam 5) & 453\\
    && OWSM-CTC (ours) & 1\\
    \bottomrule
  \end{tabular}
  }
  \vskip -0.05in
  \caption{Comparison of the number of decoding failures in all ST test sets. There are 286k samples in total.
  }
  \label{tab:robustness-failures}
\end{table}
\endgroup

\section{Conclusion}
We propose OWSM-CTC, a novel encoder-only speech foundation model built upon 180k hours of public audio data and open-source toolkits. OWSM-CTC employs multi-task self-conditioned CTC for multilingual ASR, any-to-any ST, and LID.
We conduct extensive experiments to compare OWSM-CTC with the encoder-decoder OWSM models trained on the same data.
We find that OWSM-CTC achieves competitive performance on ASR and superior performance on ST for both X-to-En (24\% relative improvement) and En-to-X (13\% relative improvement), while being more robust and 3 to 4 times faster at inference time.
Additionally, OWSM-CTC improves the long-form ASR WER with 20 times faster inference due to the batched parallel decoding.
OWSM-CTC also outperforms the baselines on LID.
To promote open research on large speech models, we will publicly release our code, pre-trained model weights and training logs.

\section*{Limitations}
Although OWSM-CTC reduces the training cost by 22\% compared to OWSM v3.1, it still requires nearly 20k GPU hours, which is nontrivial.
OWSM-CTC can generate incorrect ASR or ST outputs due to limited training data in certain languages.
Care should be taken when using our model for low-resource ASR or ST.
Besides, we have only evaluated our model with greedy decoding as it has the fastest inference speed. The non-autoregressive model sometimes makes mistakes in spelling or grammar due to a lack of language models.

\section*{Broader Impacts and Ethics}
Our OWSM-CTC is a novel encoder-only speech foundation model built upon public datasets and open-source toolkits. Compared to other popular choices, it achieves very strong performance and efficiency.
We adhere to the ACL ethics policy and there is no violation of privacy in our experiments.
We plan to publicly release all scripts, pre-trained models, and training logs, which can promote transparency and open science. We believe this will benefit the entire speech research community and it can make the latest speech technology available to a broader range of people all over the world.

\section*{Acknowledgements}
We want to thank Amazon AGI for funding.
Our computing resources are supported by PSC Bridges2 and NCSA Delta via ACCESS allocation CIS210014, under National Science Foundation grants \#2138259, \#2138286, \#2138307, \#2137603, and \#2138296.

\bibliography{anthology,custom}

\appendix

\section{Details of Experimental Setups}
\label{app:details}

\begingroup
\begin{table*}[tb]
  \centering
  \resizebox {0.95\textwidth} {!} {
  \begin{tabular}{lcccccc}
    \toprule
    Model & Unlabeled & English ASR & Other ASR & ST & Languages & Vocabulary Size \\
    \midrule
    \multicolumn{7}{l}{\textbf{Whisper}~\citep{whisper}}\\
    \quad Initial versions & - & 438k hours & 117k hours & 125k hours & 99 & 52k\\
    \quad large-v3 & 4M hours & \multicolumn{3}{c}{1M hours of labeled in total} & 100 & 52k\\
    \midrule
    \multicolumn{7}{l}{\textbf{OWSM v3.1}~\citep{owsm-v31}}\\
     & - & 73k hours & 67k hours & 40k hours & 151 & 50k\\
    \midrule
    \multicolumn{7}{l}{\textbf{OWSM-CTC (ours)}}\\
     & - & 73k hours & 67k hours & 40k hours & 151 & 50k\\
    \bottomrule
  \end{tabular}
  }
  \caption{Details of training data. Our data is prepared using the scripts released by OWSM v3.1~\citep{owsm-v31}.}
  \label{tab:training-data}
\end{table*}
\endgroup

\begingroup
\begin{table*}[tb!]
  \centering
  \resizebox {\textwidth} {!} {
  \begin{tabular}{lccccccc}
    \toprule
    Model & Params & Encoder & Decoder & Layers & Hidden Size & Attention Heads & Time Shift\\
    \midrule
    \multicolumn{8}{l}{\textbf{Whisper}~\citep{whisper}}\\
    \quad tiny & 39M & Transformer & Transformer & 4 & 384 & 6 & 20ms\\
    \quad base & 74M & Transformer & Transformer & 6 & 512 & 8 & 20ms\\
    \quad small & 244M & Transformer & Transformer & 12 & 768 & 12 & 20ms\\
    \quad medium & 769M & Transformer & Transformer & 24 & 1024 & 16 & 20ms\\
    \quad large & 1.55B & Transformer & Transformer & 32 & 1280 & 20 & 20ms\\
    \quad large-v3 & 1.55B & Transformer & Transformer & 32 & 1280 & 20 & 20ms\\
    \midrule
    \multicolumn{8}{l}{\textbf{OWSM v3.1}~\citep{owsm-v31}}\\
    \quad base & 101M & E-Branchformer & Transformer & 6 & 384 & 6 & 40ms\\
    \quad medium & 1.02B & E-Branchformer & Transformer & 18 & 1024 & 16 & 40ms\\
    \midrule
    \multicolumn{8}{l}{\textbf{OWSM-CTC (ours)}}\\
    \quad medium & 1.01B & E-Branchformer & - & 27 & 1024 & 16 & 80ms\\
    \bottomrule
  \end{tabular}
  }
  \caption{Details of model architectures. Whisper~\citep{whisper} and OWSM v3.1~\citep{owsm-v31} are encoder-decoder models, whereas our OWSM-CTC is an encoder-only model. We mostly follow the design of OWSM v3.1 medium, but we increase the number of encoder layers to match the overall model size.}
  \label{tab:model-arch}
\end{table*}
\endgroup

\begingroup
\begin{table*}[tb]
  \centering
  \resizebox {0.9\textwidth} {!} {
  \begin{tabular}{lccccc}
    \toprule
    Model & Batch Size & Total Steps & Warmup Steps & Max Learning Rate & InterCTC Layers $\Sc$\\
    \midrule
    \multicolumn{6}{l}{\textbf{OWSM v3.1}~\citep{owsm-v31}}\\
    \quad base & 256 & 675k & 60k & 1e-3 & -\\
    \quad medium & 256 & 675k & 60k & 2e-4 & -\\
    \midrule
    \multicolumn{6}{l}{\textbf{OWSM-CTC (ours)}}\\
    \quad medium & 256 & 600k & 60k & 2e-4 & 6, 12, 15, 21\\
    \bottomrule
  \end{tabular}
  }
  \caption{Training hyperparameters. We mostly follow the training setups of OWSM v3.1 medium~\citep{owsm-v31}. As described in Section~\ref{subsec:speech-encoder}, we employ self-conditioned CTC at four intermediate layers. 
  }
  \label{tab:training-hyper}
\end{table*}
\endgroup

\begingroup
\begin{table*}[tb]
  \centering
  \resizebox {\textwidth} {!} {
  \begin{tabular}{lccccc}
    \toprule
    Downsampling Strategy & Params & GPU VRAM ($\downarrow$) & Speed-up ($\uparrow$) & ASR WER ($\downarrow$) & ST BLEU ($\uparrow$) \\
    \midrule
    4x in CNN & 55M & 38GB & 1.00x & \textbf{8.3} & \textbf{22.0}\\
    6x in CNN & 55M & 22GB & 1.12x & 8.6 & 21.3\\
    8x in CNN & 55M & \textbf{19GB} & \textbf{1.13x} & 8.8 & 21.5\\
    4x in CNN + 2x in the middle of Encoder & 55M & 38GB & 1.03x & 9.7 & 21.6\\
    \bottomrule
  \end{tabular}
  }
  \caption{Comparison of different downsampling strategies on MuST-C v2 En-De. The other configurations, such as batch size, are kept the same. Using 4x downsampling achieves the best ASR and ST results, while using 8x downsampling significantly reduces the GPU memory usage, which enables a larger batch size per GPU. We employ 8x downsampling in our large-scale OWSM-CTC to reduce training costs.}
  \label{tab:ablation-downsampling}
\end{table*}
\endgroup

\begingroup
\setlength{\tabcolsep}{8pt}
\begin{table*}[t!]
  \centering
  \resizebox {0.85\textwidth} {!} {
  \begin{tabular}{lrcc}
    \toprule
    ASR-Only CTC Layers & Task-Dependent CTC Layers & ASR WER ($\downarrow$) & ST BLEU ($\uparrow$) \\
    \midrule
    - & 6, 12, 18, 24 & \multicolumn{2}{c}{diverged}\\
    6 & 12, 18, 24 & 9.0 & \textbf{21.6}\\
    6, 12 & 18, 24 & 8.8 & 21.5\\
    6, 12, 18 & 24 & \textbf{8.4} & 21.2\\
    \bottomrule
  \end{tabular}
  }
  \caption{Effect of the CTC type. This small-scale model has 24 layers with 8x downsampling in CNN. As described in Section~\ref{subsec:speech-encoder}, we employ self-conditioned CTC at some intermediate layers. These CTC layers can perform a single task like ASR or multiple tasks depending on the task specifier. If we allow all CTC layers to perform multiple tasks (ASR and ST), the model cannot converge from scratch. Therefore, we leverage the first few CTC layers for ASR only and the remaining ones for multi-tasking.}
  \label{tab:ablation-ctc-ref}
\end{table*}
\endgroup

\begingroup
\setlength{\tabcolsep}{15pt}
\begin{table*}[tb]
  \centering
  \resizebox {0.8\linewidth} {!} {
  \begin{tabular}{l|cccc|cc}
    \toprule
    Src Lang. & de & es & fr & ca & Average ($\uparrow$) & Speed-up ($\uparrow$) \\
    \midrule
    \midrule
    data size & 4.3 & 6.7 & 4.5 & 0.2 \\
    \midrule
    \multicolumn{7}{l}{\textbf{Whisper (encoder-decoder)}~\citep{whisper}}\\
    base & 11.4 & 19.2 & 13.1 & 9.7 & 13.4 & 1.84x\\
    small & 25.0 & 32.8 & 26.4 & 21.7 & 26.5 & 1.54x\\
    medium & \textbf{33.6} & \textbf{39.7} & \textbf{34.4} & \textbf{29.2} & \textbf{34.2} & 0.84x\\
    \midrule
    \midrule
    data size & 0.2 & 0.1 & 0.3 & 0.1 \\
    \midrule
    \multicolumn{7}{l}{\textbf{OWSM v3.1 (encoder-decoder)}~\citep{owsm-v31}}\\
    base & 7.3 & 10.0 & 11.1 & 9.0 & 9.4 & 2.78x\\
    medium & 17.1 & 22.3 & 22.7 & 18.4 & 20.1 & 1.00x\\
    \midrule
    \multicolumn{7}{l}{\textbf{OWSM-CTC (ours)}}\\
    medium & \underline{21.1} & \underline{28.2} & \underline{27.7} & \underline{23.7} & \underline{25.2} & \underline{\textbf{3.35x}}\\
    \bottomrule
  \end{tabular}
  }
  \vskip -0.05in
  \caption{BLEU ($\uparrow$) of X-to-En ST on CoVoST-2 using lowercase without punctuation. Data sizes are in thousand hours. \textbf{Bold}: the best result. \underline{Underlined}: our OWSM-CTC outperforms OWSM v3.1 medium.
  }
  \label{tab:lcrm-st-x-to-en}
\end{table*}
\endgroup

\begingroup
\setlength{\tabcolsep}{3.5pt}
\begin{table*}[tb]
  \centering
  \resizebox {\linewidth} {!} {
  \begin{tabular}{l|ccccccccccccccc|cc}
    \toprule
    Tgt Lang. & de & ca & zh & fa & et & mn & tr & ar & sv & lv & sl & ta & ja & id & cy & Average ($\uparrow$) & Speed-up ($\uparrow$) \\
    \midrule
    data size & 14.0 & 0.4 & 13.7 & 0.8 & 0.4 & 0.4 & 0.9 & 0.9 & 0.4 & 0.4 & 0.4 & 0.4 & 1.0 & 0.4 & 0.4\\
    \midrule
    \multicolumn{18}{l}{\textbf{OWSM v3.1 (encoder-decoder)}~\citep{owsm-v31}}\\
    base & 14.6 & 7.7 & 14.5 & 3.0 & 1.8 & 1.0 & 1.2 & 1.6 & 8.1 & 1.3 & 0.7 & 0.0 & 8.7 & 5.1 & 4.5 & 4.9 & 2.39x\\
    medium & 25.4 & 19.6 & 32.1 & \textbf{10.1} & 7.7 & 4.6 & 6.5 & 7.2 & 20.3 & 6.4 & 9.0 & 0.0 & \textbf{19.6} & 16.1 & 15.3 & 13.3 & 1.00x\\
    \midrule
    \multicolumn{18}{l}{\textbf{OWSM-CTC (ours)}}\\
    medium & \underline{\textbf{25.5}} & \underline{\textbf{23.0}} & \underline{\textbf{35.1}} & 10.0 & \underline{\textbf{9.2}} & \underline{\textbf{4.8}} & \underline{\textbf{6.8}} & \underline{\textbf{8.2}} & \underline{\textbf{23.8}} & \underline{\textbf{7.7}} & \underline{\textbf{12.0}} & 0.0 & 18.5 & \underline{\textbf{21.0}} & \underline{\textbf{19.4}} & \underline{\textbf{15.0}} & \underline{\textbf{4.20x}}\\
    \bottomrule
  \end{tabular}
  }
  \vskip -0.05in
  \caption{BLEU ($\uparrow$) of En-to-X ST on CoVoST-2 using lowercase without punctuation. Data sizes are in thousand hours. \textbf{Bold}: the best result. \underline{Underlined}: our OWSM-CTC outperforms OWSM v3.1 medium. Note that Whisper does not support En-to-X translation.}
  \label{tab:lcrm-st-en-to-x}
\end{table*}
\endgroup

\begingroup
\begin{table*}[tb]
  \centering
  \resizebox {\linewidth} {!} {
  \begin{tabular}{p{0.24\linewidth}p{0.24\linewidth}p{0.24\linewidth}p{0.24\linewidth}}
    \toprule
    Input audio content & Previous sentence & ASR w/o previous & ASR w/ previous\\
    \midrule
    future 's over here wind sun a new energy grid new investments to create high paying jobs repower america it 's time to get real there is an old african proverb that says if you want to go quickly go alone if you want to go far go together we need to go far quickly thank you very much 
    & with one hundred percent clean electricity within ten years a plan to put america back to work make us more secure and help stop global warming finally a solution that 's big enough to solve our problems repower america find out more this is the last one it 's about repowering america one of the fastest ways to cut our dependence on old dirty fuels that are killing our planet
    & Future's over here. Wind, sun. A new energy grid. New investments to create high-pan jobs. Repower America. It's time to get real. There's an old African proverb that says, "If you want to go quickly, go alone. if you want to go far, go together." We need to go far quickly. Thank you very much. (Applause)
    & future 's over here wind sun a new energy grid new investments to create high pan jobsrepower america it 's time to get real there 's an old african proverb that says if you want to go quickly go alone if you want to go far go together we need to go far quickly thank you very much\\
    \bottomrule
  \end{tabular}
  }
  \caption{Using a previous sentence as the prompt might change the output style.
  The optional prompt encoder is defined in \autoref{fig:model-arch} and Section~\ref{subsec:prompt-encoder}.
  }
  \label{tab:example-prompt}
\end{table*}
\endgroup

\begingroup
\begin{table*}[tb]
  \centering
  \resizebox {\linewidth} {!} {
  \begin{tabular}{p{0.4\linewidth}p{0.4\linewidth}p{0.4\linewidth}}
    \toprule
    \textbf{Groundtruth reference} & \textbf{OWSM v3.1 output} & \textbf{OWSM-CTC output (ours)}\\
    \midrule
    in search of the mythical treasure your grandfather is supposed to have secreted there he laughed and the girl instinctively shuddered with a newborn distrust there was no mirth in the sound
    & in search of the mythical treasure your grandfather is supposed to have secreted there \textcolor{red}{ha ha ha ha ha ha ha ha ha ha ha ha ha ha ha ha ha ha ... }
    & in search of the mythical treasure your grandfather is supposed to have secreted there he laughed and the girl instinctively shuddered with a new-born distrust there was no mirth in the sound\\
    \midrule
    and with her they began a national tour that took them all around the country & they take a national gira which leads to \textcolor{red}{rerererererererererererererere ...} & with learn a national tour that leads them to run the entire country\\
    \bottomrule
  \end{tabular}
  }
  \caption{Autoregressive decoding sometimes gets trapped in a loop in both ASR (row 1, MLS En) and ST (row 2, CoVoST-2 Es-En). Our OWSM-CTC is more robust.
  }
  \label{tab:robustness-example}
\end{table*}
\endgroup

\subsection{Training data}
\label{app:training-data}
\autoref{tab:training-data} summarizes the training data statistics. We prepare the training data mixture using the scripts publicly released by OWSM v3.1~\citep{owsm-v31}. This ensures a fair comparison between our OWSM-CTC and the previously released encoder-decoder OWSM models.

Our use of the data is consistent with their intended use. 
These datasets have been widely used in speech research. They do not violate the privacy of creators or users, nor do they contain any offensive content.
Specifically, the individual training datasets and licenses are listed below:
AIDATATANG (CC BY-NC-ND 4.0)\footnote{\url{https://www.openslr.org/62/}}, AISHELL-1 (Apache 2.0)~\cite{aishell}, AMI (CC BY 4.0)~\cite{ami}, Babel\footnote{\url{https://www.iarpa.gov/research-programs/babel}}, CommonVoice (CC0-1.0)~\cite{commonvoice}, CoVoST2 (CC BY-NC 4.0)~\cite{covost2}, Fisher Switchboard (LDC)~\cite{swbd}, Fisher Callhome Spanish (LDC)~\cite{post-etal-2013-improved}, FLEURS (CC-BY-4.0)~\cite{fleurs}, Googlei18n\footnote{Resources 32, 35, 36, 37, 41, 42, 43, 44, 52, 53, 54, 61, 63, 64, 65, 66, 69, 70, 71, 72, 73, 74, 75, 76, 77, 78, 79, and 86 from \url{openslr.org}.}, GigaSpeech (Apache 2.0)~\cite{gigaspeech}, GigaST (CC BY-NC 4.0)~\cite{gigast}, KsponSpeech (MIT License)~\cite{ksponspeech}, LibriSpeech (CC BY 4.0)~\cite{librispeech}, Multilingual LibriSpeech (CC BY 4.0)~\cite{mls}, MagicData (CC BY-NC-ND 4.0)\footnote{\url{https://openslr.org/68/}}, MuST-C (CC BY NC ND 4.0 International)~\cite{mustc}, SPGISpeech~\cite{spgispeech}, TEDLIUM3 (CC BY-NC-ND 3.0)~\cite{tedlium3}, ReazonSpeech (Apache 2.0 / CDLA-Sharing-1.0)~\cite{reazonspeech}, Russian OpenSTT (CC-BY-NC)\footnote{\url{https://github.com/snakers4/open_stt}}, VCTK (CC BY 4.0)\footnote{\url{https://huggingface.co/datasets/vctk}}, VoxForge (GPL)\footnote{\url{https://www.voxforge.org/}}, VoxPopuli (Attribution-NonCommercial 4.0 International)~\cite{wang-etal-2021-voxpopuli}, WenetSpeech (Creative Commons Attribution 4.0 International License)~\cite{wenetspeech}.

\subsection{Model architectures}
\label{app:model-arch}
\autoref{tab:model-arch} shows the model configurations. Our OWSM-CTC mostly follows the design of OWSM v3.1 medium~\citep{owsm-v31}, but we only use an encoder. To match the total model size, we increase the number of layers to 27, leading to a total of 1B parameters. 
Note that the sequence length of the encoder is usually longer than that of the decoder. Hence, the encoder-only model can have a higher computational cost than the encoder-decoder model.
To alleviate this issue, we apply a larger downsampling rate in the CNN module to reduce the sequence length. Our final time shift is 80ms, as opposed to 40ms of the encoder-decoder OWSM models. We observe that our training time for a fixed number of updates is roughly the same as that of OWSM v3.1 medium. We also investigated different downsampling strategies at a smaller scale, as discussed in Appendix~\ref{app:effect-downsampling} and \autoref{tab:ablation-downsampling}.

\subsection{Training hyperparameters}
\label{app:training-hyper}
\autoref{tab:training-hyper} presents the training hyperparameters of OWSM v3.1 and our OWSM-CTC. Again, we follow the previous OWSM v3.1~\citep{owsm-v31} for a fair comparison, except that we adopt self-conditioned CTC~\citep{self-conditioned-ctc} at four intermediate layers (see Section~\ref{subsec:speech-encoder}).

\section{Small-Scale Ablation Studies}

Before the large-scale training using the entire 180k hours of audio data, we conducted preliminary experiments on MuST-C v2 En-De~\citep{mustc} to investigate the effect of the CNN downsampling rate and the choice of the task for intermediate CTC layers. Specifically, we train 24-layer E-Branchformer-CTC models on the combined ASR and ST data from MuST-C v2 En-De. The input is always English audio, but the output can be the English ASR transcript or its German translation depending on the task specifier (see \autoref{fig:model-arch}).

\subsection{Effect of downsampling strategies}
\label{app:effect-downsampling}

\autoref{tab:ablation-downsampling} compares different downsampling strategies while the other configurations are kept the same. The attention is implemented with FlashAttention~\citep{flash-attn}. Self-conditioned CTC is applied at three intermediate layers: 6, 12, and 18. The first two CTC layers always perform ASR, while the others are task-dependent. The results show that using 8x downsampling in the CNN module leads to a slight degradation on WER and BLEU but reduces the GPU memory usage by half. We thus decide to employ 8x downsampling in our large-scale OWSM-CTC, enabling a doubled batch size per GPU. As mentioned in Appendix~\ref{app:model-arch}, with this strategy, we observe a similar training speed compared to the encoder-decoder OWSM model.

\subsection{Choice of the CTC task}
\label{app:choice-ctc-ref}
As discussed in Section~\ref{subsec:speech-encoder}, the intermediate CTC layers can be configured to perform a specific task like ASR or multiple tasks depending on the input task token.
\autoref{tab:ablation-ctc-ref} compares different choices at a small scale using MuST-C v2 En-De.
If all CTC layers are task-dependent (i.e., multi-tasking), the model cannot converge when trained from scratch.
As more layers are used for ASR only, the ASR WER improves, but the ST BLEU decreases slightly. A good trade-off is to use the first half for ASR only and the second half for multi-tasking.
Therefore, in our large-scale OWSM-CTC with 27 layers, we configure the 6th, 12th, and 15th layers to perform ASR only and the other two CTC layers (i.e., 21st and 27th layers) to be multi-tasking. This design also mimics the conventional cascaded system for ST.

\section{More Results of ST}
\label{app:more-st-results}

Section~\ref{subsec:st-results} shows the BLEU scores using true case with punctuation. In this section, \autoref{tab:lcrm-st-x-to-en} and \autoref{tab:lcrm-st-en-to-x} present BLEU in lowercase without punctuation, which is consistent with the setup in prior work~\citep{owsm-v31}.
The findings are very consistent with those in Section~\ref{subsec:st-results}. Our OWSM-CTC achieves higher BLEU scores with faster inference speeds than the encoder-decoder OWSM v3.1 in general.

\section{Effect of text prompt}
\label{app:example-prompt}

\autoref{tab:example-prompt} presents an example from TEDLIUM, where the text prompt changes the output style. 
When there is no prompt, the ASR output of OWSM-CTC is in true case with punctuation, and the apostrophes are combined with the previous words. However, when the previous sentence is used as a prompt, the style of the ASR hypothesis becomes more similar to that of the prompt. Specifically, the text is now in lowercase without punctuation marks, and the apostrophes are separate from previous words. This style is closer to the groundtruth transcript.

Although the above example looks promising for biasing the model's output toward certain directions, we note that this is not guaranteed to work in a zero-shot manner. We have also tried a few examples for zero-shot contextual biasing, where we provide a few biasing words in the prompt (e.g., person names), but we find that the model may not generate the correct word in many cases. This is mainly because the model is not really trained to perform this type of task - we just provide the previous sentence (according to some probability) as the prompt during training, which might not be useful at all; thus, the non-autoregressive model can simply ignore it in most cases.
A more practical way to utilize this feature is to fine-tune our pre-trained model using some carefully designed data for contextual biasing. We will explore this in the future.

\section{Robustness}
\label{app:robustness}

\autoref{tab:robustness-example} shows that autoregressive decoding sometimes fails to generate the correct output for either ASR or ST, while non-autoregressive decoding is generally more robust to this type of error.

\end{document}